\documentclass[letterpaper, 10 pt, conference]{ieeeconf}
\usepackage[bookmarks=false,hidelinks=true]{hyperref}
\usepackage{graphicx,dblfloatfix}
\usepackage{amsmath}
\usepackage{amssymb}
 
\usepackage{float}
\usepackage{listings}
\usepackage{color}
\usepackage{xcolor}
\usepackage{caption}
\usepackage{subfigure}
\usepackage{amsmath}
\usepackage{algorithm}
\usepackage{algpseudocode}
\usepackage{epigraph}
\usepackage{booktabs}
\usepackage{setspace}
\usepackage{amsfonts,amssymb}

\IEEEoverridecommandlockouts
\title{\LARGE \bf Dense RGB-D semantic mapping with\\
 Pixel-Voxel neural network}

\author{Cheng Zhao$^{1,2}$, Li Sun$^{1}$, Pulak Purkait$^{2}$ and Rustam Stolkin$^{1}$
\thanks{$^{1}$ Extreme Robotics Lab, University of Birmingham, Birmingham, UK, B15 2TT.
        {\tt\small IRobotCheng@gmail.com}. $^{2}$Cambridge Research Lab, Toshiba Research Europe, Cambridge, UK, CB4 0GZ.}%
}
       
\begin{document}              
\maketitle
\thispagestyle{empty}
\pagestyle{empty}
\begin{abstract}
For intelligent robotics applications, extending 3D mapping to 3D semantic mapping enables robots to, not only localize themselves with respect to the scene's geometrical features but also simultaneously understand the higher level meaning of the scene contexts. Most previous methods focus on geometric 3D reconstruction and scene understanding independently notwithstanding the fact that joint estimation can boost the accuracy of the semantic mapping. In this paper, a dense RGB-D semantic mapping system with a Pixel-Voxel network is proposed, which can perform dense 3D mapping while simultaneously recognizing and semantically labelling each point in the 3D map. The proposed Pixel-Voxel network obtains global context information by using PixelNet to exploit the RGB image and meanwhile, preserves accurate local shape information by using VoxelNet to exploit the corresponding 3D point cloud. Unlike the existing architecture that fuses score maps from different models with equal weights, we proposed a Softmax weighted fusion stack that adaptively learns the varying contributions of PixelNet and VoxelNet, and fuses the score maps of the two models according to their respective confidence levels. The proposed Pixel-Voxel network achieves the state-of-the-art semantic segmentation performance on the SUN RGB-D benchmark dataset. The runtime of the proposed system can be boosted to 11-12Hz, enabling near to real-time performance using an i7 8-cores PC with Titan X GPU.
\end{abstract}
\section{Introduction}\label{sec:1}
A Real-time 3D semantic mapping is desired in a lot of robotics applications, such as autonomous navigation and robot arm manipulation. The inclusion of semantic information with a 3D dense map is much useful than geometric information alone in robot-human or robot-environment interaction. It enables robots to perform advantage tasks like "nuclear wastes classification and sorting" or "autonomous warehouse package delivery" more intelligently.  

A variety of well-known methods such as RGB-D SLAM~\cite{endres20143}, Kinect Fusion~\cite{newcombe2011kinectfusion} and ElasticFusion~\cite{whelan2015elasticfusion} can generate dense or semi-dense 3D map from RGB-D videos. But those 3D maps contain no semantic-level understanding of the observed scenes. Meanwhile, the semantic segmentation achieved a significant progress with advantage of convolution neural network. Thus far, FCN~\cite{long2015fully}, SegNet~\cite{badrinarayanan2015segnet} and Deeplab~\cite{chen2016deeplab} are the most popular methods for RGB level semantic segmentation. FuseNet~\cite{hazirbas2016fusenet} and LSTM-CF~\cite{li2016lstm} take advantage of both RGB and depth images to improve semantic segmentation. PointNet~\cite{qi2016pointnet} is the forerunner for 3D semantic segmentation that consumes an unordered point cloud. 

\begin{figure}[thpb]
	\centering
	\includegraphics[width= 0.45\textwidth]{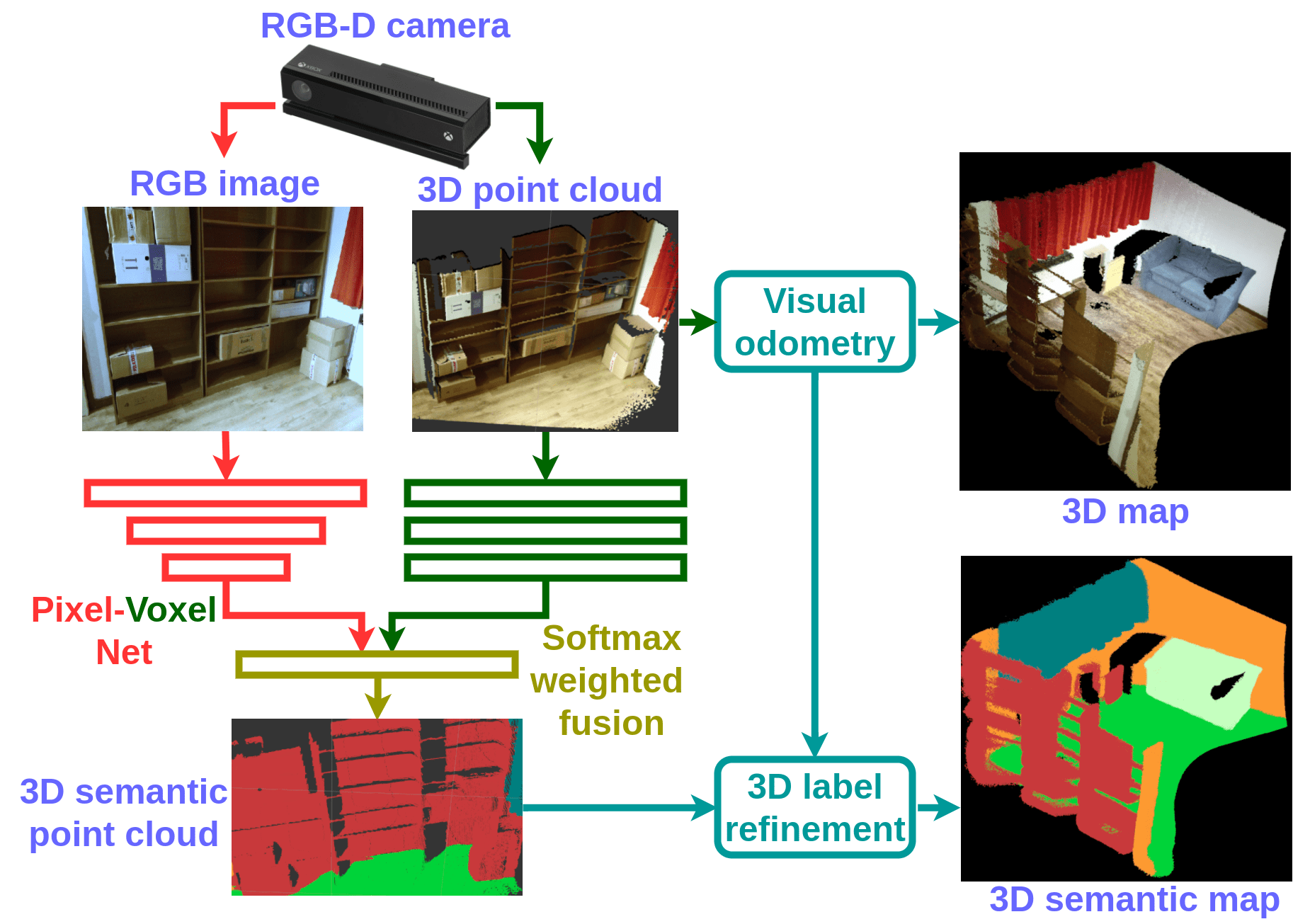}
	\caption{\textbf{The pipeline of dense RGB-D semantic mapping with Pixel-Voxel neural network.} The RGB image and point cloud are obtained directly from an RGB-D camera Kinect V2. The RGB and point cloud data-pair of each key-frame is fed into the Pixel-Voxel network for semantic segmentation. Then the semantically labelled point clouds are combined incrementally through the visual odometry of RGB-D SLAM. Meanwhile, the label probability of each voxel is refined by a recursive Bayesian update. Finally, the dense 3D semantic map is generated.}
	\label{fig:Pipeline}
\end{figure}

During RGB-D mapping, both RGB image with rich contextual information and point cloud with rich 3D geometric information can be obtained directly. To date, there are no existing methods that make use of both RGB and point cloud for the semantic segmentation and mapping. In this paper, we proposed a dense RGB-D semantic mapping system with a Pixel-Voxel neural network which can perform dense 3D mapping while simultaneously recognizing and semantically labelling each point in the 3D map. The main contributions of this paper can be summarized as follows: 

\begin{itemize}

\item A Pixel-Voxel network consuming RGB image and point cloud is proposed, which can obtain global context information through PixelNet and meanwhile, preserve accurate local shape information through VoxelNet. This mutual promotion model achieves the state-of-the-art semantic segmentation performance on SUN RGB-D\footnote{\url{http://rgbd.cs.princeton.edu/challenge.html}} dataset.    

\item A Softmax weighted fusion stack is proposed to adaptively learn the varying contribution of different models. It can fuse the score maps from different models according to their respective confidence levels. The number of input models for fusion can be arbitrary. This stack can be inserted to any kind of network to perform fusion style end-to-end learning.

\item A dense 3D semantic mapping system integrating Pixel-Voxel network with RGB-D SLAM is developed. Its runtime can be boosted to $11-12Hz$ using an i7 8-cores PC with Titan X GPU, which can nearly satisfy the requirement of real-time applications.

\end{itemize}

The rest of this paper is organized as follows. The related work is reviewed in Section \ref{sec:2} firstly. Then the details of the proposed methods are introduced in Section \ref{sec:3}. The experimental results and analyses are given in Section \ref{sec:4}. Finally, we conclude the paper in Section \ref{sec:5}.

\section{Related work}\label{sec:2}
The existing works are categorized and described in the following two subsection, dense 3D semantic mapping in Section \ref{sec:2.1} and semantic segmentation in Section \ref{sec:2.2}, followed by a discussion in Section \ref{sec:2.3}

\subsection{Dense 3D semantic mapping}\label{sec:2.1}
To the best of our knowledge, the online dense 3D semantic mapping can be further grouped into three main sub-categories: semantic mapping based on 3D template matching~\cite{salas2013slam++}\cite{tateno20162}, 2D/2.5D semantic segmentation~\cite{hermans2014dense}\cite{vineet2015incremental}\cite{mccormac2017semanticfusion}\cite{tateno2017cnn}\cite{zhao2017fully} and RGB-D data association from multiple viewpoints~\cite{xiang2017rnn}\cite{ma2017multi}\cite{mustafa2017semantically}.  

The first kind of methods such as SLAM++~\cite{salas2013slam++} can only recognise the known 3D objects in a pre-defined database. It is limited to can only be used in the situations where many repeated and identical objects are present for semantic mapping. 

For the second kind of methods, both~\cite{hermans2014dense} and~\cite{vineet2015incremental} adopt human-design features with Random Decision Forests to perform per-pixel label predictions of the incoming RGB videos. Then all the semantically labelled images are associated together using a visual odometry to generate the semantic map. Because of the state-of-the-art performance provided by the CNN-based scene understanding, SemanticFusion~\cite{mccormac2017semanticfusion} integrates deconvolution neural networks~\cite{noh2015learning} with ElasticFusion~\cite{whelan2015elasticfusion} to a real-time capable ($25Hz$) semantic mapping system. All of those three methods require fully connected CRF~\cite{krahenbuhl2011efficient} optimization as an offline post-processing, i.e., the best performance semantic mapping is not an online system. Zhao \emph{et al.}~\cite{zhao2017fully}. proposed the first system to perform simultaneous 3D mapping and pixel-wise material recognition. It integrates CRF-RNN~\cite{zheng2015conditional} with RGB-D SLAM~\cite{endres20143} and the post-processing optimization is not required. Keisuke \emph{et al.}~\cite{tateno2017cnn} proposed a real-time dense monocular CNN-SLAM, which can perform depth prediction and semantic segmentation simultaneously from a single image using a deep neural network.

All the above methods mainly focus on semantic segmentation using a single image and they only perform 3D label refinement through a recursive Bayesian update using a sequence of images. However, they do not take full advantage of the associated information provided by multiple viewpoints of a scene. Yu \emph{et al.}~\cite{xiang2017rnn} proposed a DA-RNN integrated with Kinect Fusion~\cite{newcombe2011kinectfusion} for 3D semantic mapping. DA-RNN employs a recurrent neural network to tightly combine the information contained in multiple viewpoints of an RGB-D video stream to improve the semantic segmentation performance. Ma \emph{et al.}~\cite{ma2017multi} proposed a multi-view consistency layer which can use multi-view context information for object-class segmentation from multiple RGB-D views. It utilizes the visual odometry trajectory from RGB-D SLAM~\cite{endres20143} to wrap semantic segmentation between two viewpoints. In addition, Armin \emph{et al.}~\cite{mustafa2017semantically} proposed a network architecture for spatially and temporally coherent semantic co-segmentation and mapping of complex dynamic scenes from multiple static or moving cameras. 

\subsection{Semantic segmentation}\label{sec:2.2}
According to the type of input data, semantic segmentation can be further grouped into three main sub-categories: RGB~\cite{long2015fully}\cite{badrinarayanan2015segnet}\cite{chen2016deeplab}\cite{zheng2015conditional}\cite{noh2015learning}, RGB-D~\cite{hazirbas2016fusenet}\cite{li2016lstm}\cite{he2017std2p} and 3D~\cite{qi2016pointnet}\cite{qi2017pointnet++} semantic segmentation.

FCN~\cite{long2015fully} is the first end-to-end fashion network instead of using hand-crafted features for semantic segmentation. It replaces the fully connected layers of the classification network with the convolution layers to output the coarse map and utilizes a skip architecture to refine it. DeconvNet~\cite{noh2015learning} composing of deconvolution and unpooling layers, utilizes the fractionally strided convolutions to alleviate the limited resolution of labelling problem. SegNet~\cite{badrinarayanan2015segnet} proposed an encoder-decoder architecture, which records the indices of max pooling for up-sampling. DeepLab~\cite{chen2016deeplab} makes use of dilated convolutions~\cite{yu2015multi} to increase the receptive field without down-sampling the feature map. CRF as RNN~\cite{zheng2015conditional} reformulates the mean-field inference in dense CRF as an RNN architecture that enables it to integrate with CNN as a fully end-to-end network.        

FuseNet~\cite{hazirbas2016fusenet} can fuse RGB and depth image cues in a single encoder-decoder CNN architecture for RGB-D semantic segmentation. LSTM-CF~\cite{li2016lstm} network fuses contextual information from multiple channels of RGB and depth image through stacking several convolution layers and a long short-term memory layer. FuseNet normalises the depth value into the interval of $[0, 255]$ to have the same spatial range as color images, while LSTM-CF network transforms depth image to HHA image to have 3 channels as the color image.  The HHA representation can improve the depth semantic segmentation, however, HHA representation requires high computational cost and hence cannot be performed in real-time. In addition, STD2P~\cite{he2017std2p} proposes a novel superpixel-based multi-view convolutional neural network for RGB-D semantic segmentation, which uses Spatio-temporal pooling layer to aggregate information over space and time.    

The forerunner work PointNet~\cite{qi2016pointnet} provides a unified architecture for both classification and segmentation which consumes the raw unordered point clouds as input. PointNet only employs a single max-pooling to generate the global feature which describes the original input clouds, thus it does not capture the local structures induced by the 3D metric space points live in. In the improved version PointNet++~\cite{qi2017pointnet++}, it proposed a hierarchical neural network. It applies PointNet recursively on a nested partitioning of the input point set, which enables it to learn local features with increasing contextual scales. 

\subsection{Discussion}\label{sec:2.3}
For the RGB semantic segmentation, CNN-based methods always struggle with the balance between global and local information. The global context information can alleviate the local ambiguities to improve the recognition performance, while local information is crucial to obtain accurate per-pixel accuracy, i.e., shape information. But after several of pooling layers, the resolution of the feature map decreases significantly. It means a lot of shape information is lost. How to increase the receptive field to get more global context information and meanwhile, preserve a high resolution of feature map is still an open problem. 3D geometric data such as point cloud which has additional dimension can provide very useful spatial information. But because of the unordered property of point cloud, the conventional pooling layer cannot be used. It is difficult to obtain the context information in different scales for the point cloud. On the other hand, the resolution of point cloud would not decrease because of the absence of conventional pool layers, i.e., it can keep the original spatial information of the data. 

Intuitively, combining RGB-based network and point cloud-based network together can alleviate each other's drawbacks and take advantage of each other's advantages. During RGB-D mapping, both RGB image and point cloud can be obtained directly from the RGB-D camera, which is easily available and enables a potential combination of the context information from RGB image and 3D shape information from the point cloud for semantic mapping. That is the main reason why a dense RGB-D semantic mapping with a Pixel-Voxel neural network is proposed in this paper.

In addition, the network in \cite{long2015fully}\cite{hazirbas2016fusenet}\cite{li2016lstm} all simply fuse the score maps from different models using equal weights. Each model should have the different contributions in different situations for different categories. So in this paper, a Softmax weighted fusion stack is proposed for adoptively learning the varying contributions of each model. 
\section{Methods}\label{sec:3}
  
\subsection{Overview}\label{sec:3.1}
\begin{figure*}[thpb]
	\centering
	\includegraphics[width= 0.8\textwidth]{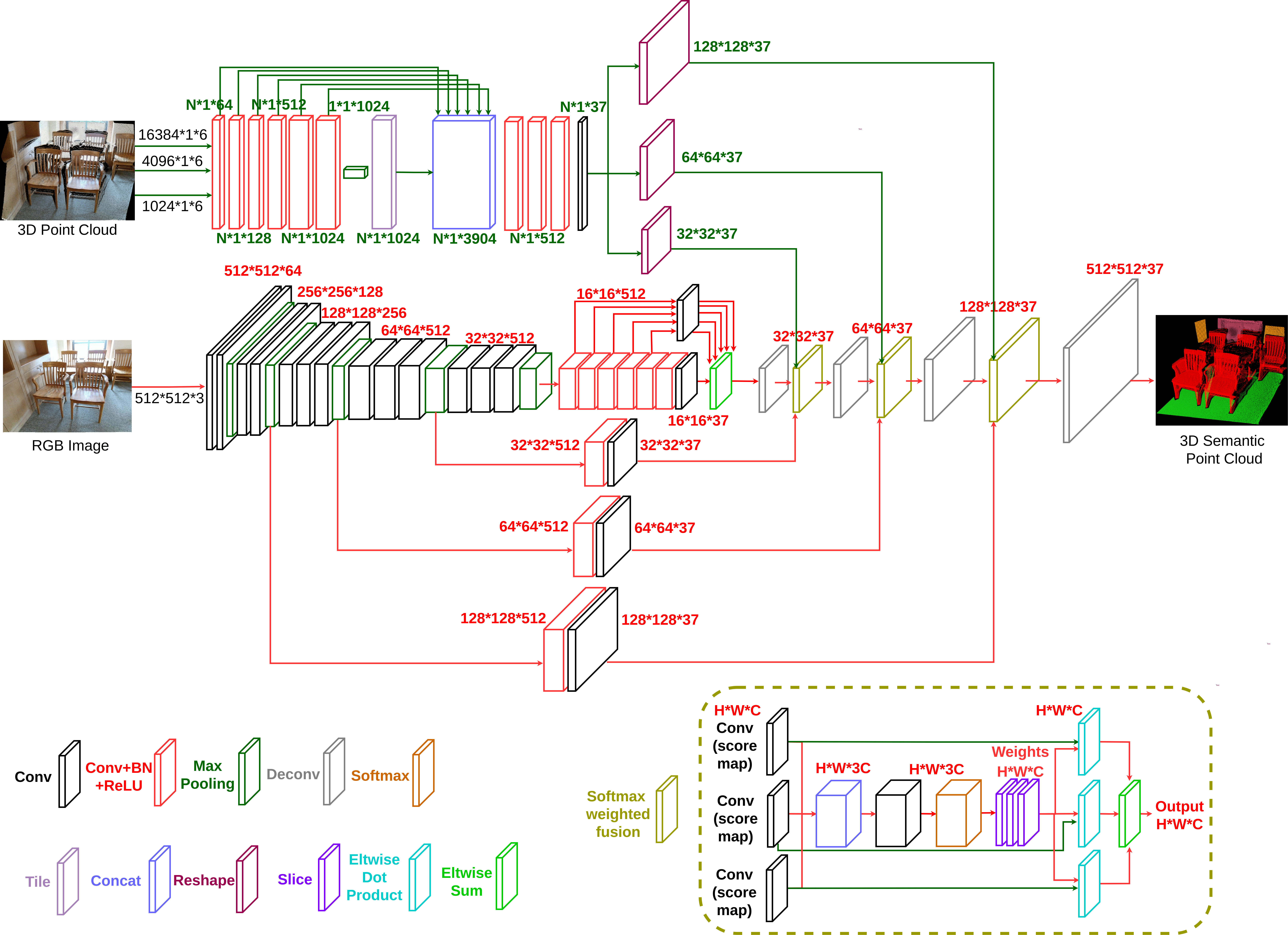}
	\caption{\textbf{The architecture of the Pixel-Voxel Network.} The PixelNet comprises three units: truncated CNN, context stack and skip architecture. The VoxelNet is composed of the convolution stacks, local and global information combination stack and reshape layer. It obtains global context information through PixelNet and meanwhile, preserves accurate local shape information through VoxelNet. The Softmax weighted fusion stack can fuse 3 score maps from PixelNet and VoxelNet together according to their respective confidence levels in different situations.}
	\label{fig:PVNet}
\end{figure*}

The pipeline of dense RGB-D semantic mapping with a Pixel-Voxel neural network is illustrated in Figure.\ref{fig:Pipeline}. The RGB image and point cloud are obtained directly from an RGB-D camera Kinect V2. The RGB and point cloud data-pair of each key-frame is fed into the Pixel-Voxel network, as shown in Figure.\ref{fig:PVNet}, for semantic segmentation. Then the semantically labelled point clouds are combined incrementally through the visual odometry of RGB-D SLAM. Meanwhile, label probability of each voxel is refined by a recursive Bayesian update. Finally, the dense 3D semantic map is generated.
 
\subsection{Pixel neural network}\label{sec:3.2}
The PixelNet is comprised of three units: truncated CNN, context stack and skip architecture. The input of PixelNet is an RGB image. For the truncated CNN, the VGG-16 or ResNet (truncated after $pool5$), pre-trained on ImageNet can be employed as the baseline. After truncated CNN, the resolution of feature map decreases $32$ times comparing with the input image, i.e., it drops significant shape information.


Inspired by~\cite{shuai2016improving}, the context stack is on the top of pre-trained truncated CNN, which is composed of chained 6 layers of $5 \times 5 \times 512$ convolution stack ($Conv+BN+ReLU$). For the VGG-16 network, the receptive field after $pool5$ and $fc6$ layers are respectively $212\times212$ and $404\times404$, which is not sufficiently large enough to cover the $512\times512$ image that we used. The receptive field of the context stack can be described as below:

\begin{equation}
{\cal RF}_{j} = {\cal RF}_{j-1} + (k_{j} - 1) \times \prod_{i=0}^{j-1}S_{i}, j \in [1,n] 
\label{eq:receptive field}
\end{equation}

Here ${\cal RF}_{0}$ and $S_{0}$ are the receptive field and stride product before the first context stack. ${\cal RF}_{j}$, $S_{j}$ and $k_{j}$ are the receptive field, stride and kernel size of the context stack $j$. $n = 6$ is the number of context stacks. The context stack can expand the receptive field progressively to cover all the elements in the current feature map(the whole original image). In addition, the score maps of all the context stacks are fused together to aggregate multi-scale context information. The spatial dimensionality of the feature maps in context stack is unchanged as before.

The skip architecture consists of 3 skip stacks($Conv+BN+ReLU+Conv(score)$) following $pool2$, $pool3$ and $pool4$ separately. In order to prevent the network training divergence, the smaller learning rate is usually adopted for the skip architecture training as mentioned in~\cite{long2015fully}. But the skip architecture in PixelNet can be trained using a bigger learning rate because batch normalization stabilizes the back-propagated error signals. The skip architecture retains the low-level feature of the RGB image. 


\subsection{Voxel neural network}\label{sec:3.3}
The input of VoxelNet is unordered point cloud which is represented as a set of 3D points $ \lbrace p_{i} \vert i = 1,2, ... n \rbrace $ stored in a $n \times 6$ long vector. $n$ is the number of points. $p_{i}$ is a feature vector containing 6 dimension information: position information $x, y, z$ in the world coordinate and color information $R, G, B$. 

\begin{equation}
\begin{split}
[{\cal F}^{1}_{global} ... {\cal F}^{n}_{global}] = {\cal T} \big( {\cal M} \big( [f_{mlp}^{k}(p_{1}) ... f_{mlp}^{k}(p_{n})] \big) \big)
\label{eq:VoxelNet1}
\end{split}
\end{equation}

Here $f_{mlp}$ is the multi-layer perception network, i.e., $Conv+BN+ReLU$. $k$ is the number of multi-layer perception network before max pooling. Its kernel size is $1 \times 1$ and each point shares the same convolution weights. Inspired by PointNet~\cite{qi2016pointnet}, we also use max pooling operation ${\cal M}$ as the invariant function. Its kernel size is $n \times 1$. This Max pooling operation can obtain the global feature from all the points. ${\cal T}$ is the tile operation which recovers the shape of feature map from $1 \times 1$ to $n \times 1$.  
The output $[{\cal F}^{1}_{global} ... {\cal F}^{n}_{global}]$ is the global feature map of the input set. They are fed to the per point feature of multi-layer perception network to concatenate the global and local information. Because only a single max pooling is adopted to generate the global feature, it drops significant context information of the input point cloud. 

\begin{equation}
\begin{split}
[{\cal F}^{1}_{concat} ... {\cal F}^{n}_{concat}] = Concat \big( [ {\cal F}^{1}_{global}... {\cal F}^{n}_{global} ],  \\
... [f_{mlp}^{i}(p_{1})... f_{mlp}^{i}(p_{n})] \big), i \in [1,k] 
\label{eq:VoxelNet2}
\end{split}
\end{equation}

Then the new per point features are extracted though multi-layer perception network using the global and local combined point features. $m$ is the last multi-layer perception network. The reshape operation ${\cal R}$ transforms the shape of score map from $n \times 1$ to $h \times w$ through back-projection according to the $x, y, z$ values and camera intrinsic parameters, so that it can be fused with the score map of PixelNet.  

\begin{equation}
{\cal F}^{1...n}_{h \times w} = {\cal R} \big( [f_{mlp}^{m}({\cal F}^{1}_{concat}) ... f_{mlp}^{m}({\cal F}^{n}_{concat})] \big)
\label{eq:VoxelNet3}
\end{equation}

The spatial dimensionality is unchanged as the input data in VoxelNet, so it can preserve all the original shape information.
   
\subsection{Softmax weighed fusion}\label{sec:3.4}
Unlike simply fusing score maps from different models using equal weights, a Softmax weighted fusion stack is designed to learn the varying contribution of each model in different situations for different categories.    

To be precise, define the score maps ${\cal F}^{1}, {\cal F}^{2} ... {\cal F}^{n} \in \mathbb{R}^{c \times h \times w}$ are generated from $n$ different models. $c$ equals the number of categories and $ h \times w$ is the shape of score map. 

\begin{equation}
{\cal F}_{fusion} = f_{conv} \big( Concat( {\cal F}^{1}, {\cal F}^{2} ... {\cal F}^{n} ), {\cal W}_{conv} \big)
\label{eq:Softmax weighed fusion1}
\end{equation}

$f_{conv}$ is the convolution operation and $ {\cal W}_{conv} \in \mathbb{R}^{n \cdot c \times n \cdot c \times 1 \times 1} $ is the weights of convolution operation. ${\cal F}_{fusion} \in \mathbb{R}^{n \cdot c \times h \times w}$ is the fusion score map. The convolution operation can learn the correlations of the multiple score maps from $n$ different models.

\begin{equation}
{\cal W}^{1}, {\cal W}^{2}... {\cal W}^{n} = Slice \big[ \dfrac{exp( {\cal F}_{fusion} )}{ \sum_{i=1}^{n \cdot c} exp( {\cal F}_{fusion}^{i} )} \big] 
\label{eq:Softmax weighed fusion2}
\end{equation}

Softmax operation normalizes the channel values of ${\cal F}_{fusion}$ into the interval of $[0,1]$. ${\cal W}^{1}, {\cal W}^{2} ... {\cal W}^{n} \in \mathbb{R}^{c \times h \times w}$ are the corresponding weights of score maps, which denotes how confidently each model can be relied on.

\begin{equation}
{\cal F}_{sum} = \sum_{j=1}^{n} {\cal F}^{j} \odot {\cal W}^{j}, \quad s.t. \quad \textstyle \sum_{j=1}^{n \cdot c } {\cal W}^{j} = \textbf{1}  
\label{eq:Softmax weighed fusion3}
\end{equation}

${\cal F}_{sum} \in \mathbb{R}^{c \times h \times w} $ is the weighted fusion score map. $\odot$ is the element-wise multiplication operation and $\textbf{1} \in \mathbb{R}^{h \times w}$. This Softmax weighted fusion stack can fuse the score maps of arbitrary number models, and it also can be inserted to any kind of network to be trained end-to-end. As shown in Figure.\ref{fig:PVNet}, it fuses 3 score maps from PixelNet and VoxelNet together according to their respective confidence levels.

\subsection{Class-weighted loss function}\label{sec:3.5}
Imbalanced class distribution is quite common in most datasets. So focusing more on the rare classes to boost their recognition accuracy can improve the average recognition performance significantly. But the overall recognition performance will decrease lightly. We adopt the class-weighted negative log-likelihood as the loss function:

\begin{equation}
loss = - \sum_{i \in {\cal S}} \log {\cal L} \big( softmax( {\cal F}_{i} ), y_{i} \big) \cdot ( \textbf{1}_{y_{i} = j} ) 2^{\lceil \log_{10} (\delta/p_{j}) \rceil}  
\label{eq:weighted loss function}
\end{equation}

Where ${\cal L}$ is the likelihood function, ${\cal S}$ is the training data, ${\cal F}_{i}$ is the final score map and $y_{i}$ refers to the training label. $\textbf{1}_{y_{i} = j}$ is a function that returns 1 if $y_{i} = j$, otherwise 0. $p_{j}$ is the occurrence frequency of class $j$ and $2^{\lceil \log_{10} (\delta/p_{j}) \rceil}$ is the weight of class $j$. $\delta$ is the threshold of frequency criteria for the rare class. $\lceil \rceil$ is the integer ceiling operation. In this way, the rare classes can be assigned a higher weight growing exponentially. The $\delta$ is set to 2.5\% following the 85\%-15\% rule in \cite{shuai2016dag}, i.e., the frequency sum of all the rare classes is 15\%. 

\subsection{RGB-D mapping}\label{sec:3.6}
RGB-D SLAM~\cite{endres20143} is employed for dense 3D mapping. Its visual odometry can provide the transformation information between two adjacent semantically labelled point clouds. It is used for generating a global semantic map and enabling incremental semantic label fusion.  

RGB-D SLAM is a graph-based SLAM system which consists of a front-end and a back-end units. The former unit processes the RGB-D data to calculate geometric relationships between key-frames through visual features based on RANSAC. The later unit registers pairs of image frames to construct a pose graph. Subsequently, G2O\footnote{\url{http://www.openslam.org/g2o}} is used for graph optimization to obtain a maximum likelihood solution for the camera trajectory. Finally, the point clouds are combined incrementally to generate a dense global 3D map.

\subsection{3D label refinement}\label{sec:3.7}
After obtaining the semantically labelled point clouds from different viewpoints, label hypotheses are fused by a recursive Bayesian update to refine the 3D semantic map. Each voxel in the semantic point cloud stores both the label value and the corresponding discrete probability. The voxels from different viewpoints can be transformed to the same coordinate through the visual odometry of RGB-D SLAM. Then the voxel's label probability distribution can be updated by the means of a recursive Bayesian update as Equation \ref{eq:Bayesian update}.
\begin{equation}
P(x = l_{i}|I_{1,...,k}) = \dfrac{1}{Z} P(x = l_{i}|I_{1,...,k-1})P(x = l_{i}|I_{k})
\label{eq:Bayesian update}
\end{equation}     
where $l_{i}$ is the label prediction, $I_{k}$ is the $k^{th}$ frame and $Z$ is the normalizing constant. It is applied to all label probabilities of each voxel to generate a proper distribution. 
\section{Experiments}\label{sec:4}
A large-scale indoor scene dataset, i.e., SUN RGB-D dataset, is adopted for the Pixel-Voxel network evaluation. It contains $5285$ synchronized RGB-D image pairs for training/validation and $5050$ synchronized RGB-D image pairs for testing. The RGB-D image pairs with different resolutions are captured by 4 different RGB-D sensors: Kinect V1, Kinect V2, Xtion and RealSense. The SUN RGB-D scene understanding challenge is to segment $37$ indoor scene classes such as the table, chair, sofa, window, door and etc. The pixel-wise annotation is available and it has extremely unbalanced class instances. As mentioned in Section \ref{sec:3.5}, the rareness frequency threshold is set to 2.5\% in the class-weighted loss function following the 85\%-15\% rule.  

\subsection{Data augmentation and preprocessing}\label{sec:4.1}
For the PixelNet training, all the RGB images are resized to the same resolution $512 \times 512$ through a bilateral filter. We randomly flip the RGB image horizontally and scale the RGB image slightly to augment the RGB training data. 

For the VoxelNet training, there is still no large-scale ready-made 3D point cloud dataset available. We generated the point cloud using the RGB-D image pairs and the corresponding camera intrinsic parameters. Similar as mentioned in~\cite{hazirbas2016fusenet}, there are $514$ training and $558$ testing RGB-D image pairs to be excluded. Because those raw depth images contain a lot of invalid values, which gives a strong wrong supervision during training. We also randomly flip the 3D point cloud horizontally to augment the point cloud training data. It is a huge computation complexity if the original point clouds are used for VoxelNet training. So we uniformly down-sample the original point cloud to sparse point cloud in 3 different scales. The number of these sparse point clouds are $16384$, $4096$ and $1024$. Please note that the input data of VoxelNet is unordered point cloud stored in a long vector.    

\subsection{Network training}\label{sec:4.2}
The whole training process can be divided into 3 stages: PixelNet training, VoxelNet training and Pixel-Voxel network training. All the networks are trained with SGD with momentum. The batch size is set to $10$, the momentum is fixed to $0.9$ and the weight decay is fixed to $0.0005$. The new parameters are randomly initialized using Gaussian distribution with variance $10^{-2}$. 

In the PixelNet training stage, the step learning policy is adopted. The learning rate is initialized to $10^{-3}$ and decreases 10 times after 15 epochs (25 epochs in total). The learning rate of newly-initialized parameters is set to 10 times higher than that of pre-trained parameters. 

In the VoxelNet training stage, the polynomial learning policy is adopted. The learning rate is initialized to $10^{-3}$, the power is set to $0.9$ and the max iteration is set to $50000$. 

In the Pixel-Voxel network training stage, we load the pre-trained PixelNet and VoxelNet models, then finetune the whole network on the synchronized RGB and point cloud data. Because there are three Softmax weighed fusion stacks in the network, 3 times fine-tuning are required. The same learning policy as VoxelNet training is adopted. The learning rate of newly-initialized parameters in each Softmax weighted fusion stack is set to 10 times higher than that of fine-tuning parameters.

\subsection{Overall performance}\label{sec:4.3}
Following \cite{long2015fully}, three standard performance metrics for semantic segmentation: pixel accuracy, mean accuracy, mean IoU are used for the Pixel-Voxel network evaluation. The three metrics are defined as below:
\begin{itemize}
\item Pixel accuracy: $\sum_{i} n_{ii} / \sum_{i} t_{i}$

\item Mean accuracy: $(1/n_{cl})\sum_{i} n_{ii} / t_{i}$
 
\item Mean IoU: $ (1/n_{cl})\sum_{i} n_{ii}/(t_{i} + \sum_{j} n_{ji} - n_{ii}) $
\end{itemize}
where $n_{cl}$ is the number of classes, $n_{ij}$ is the number of pixels of class $i$ classified as class $j$, and $ t_{i} = \sum_{j} n_{ij} $ is the total number of pixels belong to class $i$.

The qualitative results of Pixel-Voxel network on the SUN RGB-D dataset are shown in Fig.\ref{fig:qualitative_PVNet}. Because of preserving 3D shape information through VoxelNet, it is can be seen that the results have accurate boundary shape such as the shape of the bed, close-stool and especially the legs of furniture. 

The comparison of overall performance and class-wise accuracy on the SUN RGB-D dataset are shown in Table \ref{table:The comparison overall performance on the SUN RGB-D dataset.} and Table \ref{table:The comparison class-wise accuracy on the SUN RGB-D dataset.}. The class-wise IoU of Pixel-Voxel network is also provided. We achieved 79.04\% overall pixel accuracy with 0.64\% improvement, 57.65\% mean accuracy with 4.25\% improvement and 44.24\% mean IoU with 1.94\% improvement over the state-of-the-art method \cite{lin2017exploring}. The improvements of class-wise accuracy are achieved on 30 classes. In addition, the method \cite{lin2017exploring} is painfully slowly because of the usage of high computational CRF optimization in different scales.

\newcommand{\sizeOfimage}{0.11}

\begin{figure*}[thpb]
\centering
  \label{fig:RGB39}{\includegraphics[width= \sizeOfimage\textwidth]{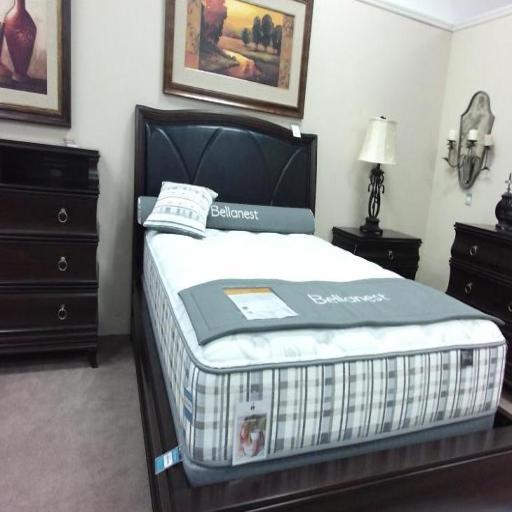}}
  \label{fig:RGB453}{\includegraphics[width= \sizeOfimage\textwidth]{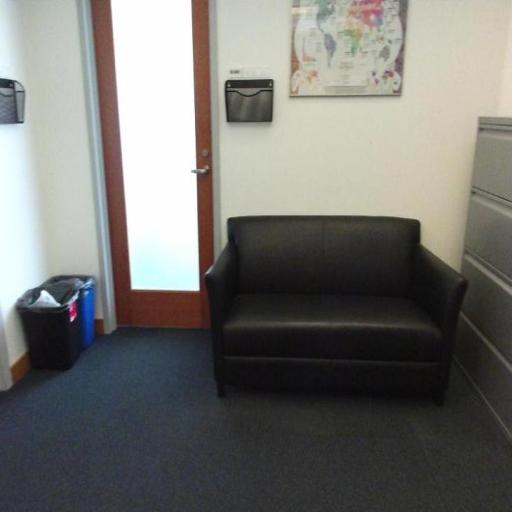}}
  \label{fig:RGB4088}{\includegraphics[width= \sizeOfimage\textwidth]{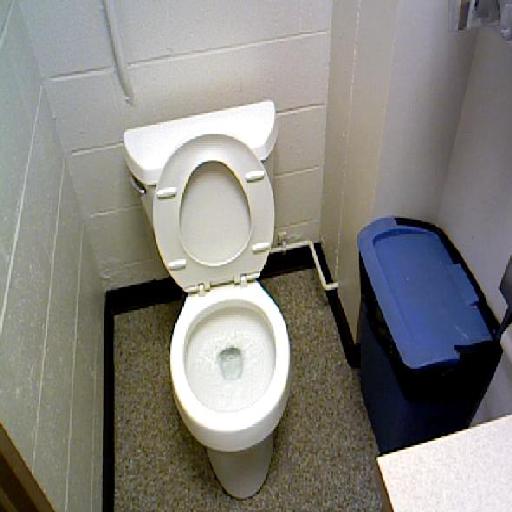}}
  \label{fig:RGB4425}{\includegraphics[width= \sizeOfimage\textwidth]{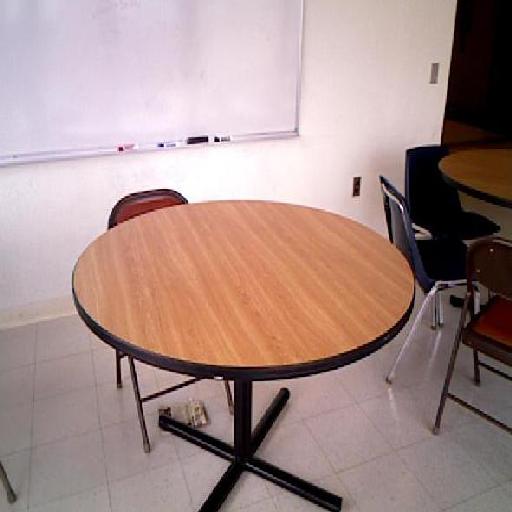}} 
  \label{fig:RGB162}{\includegraphics[width= \sizeOfimage\textwidth]{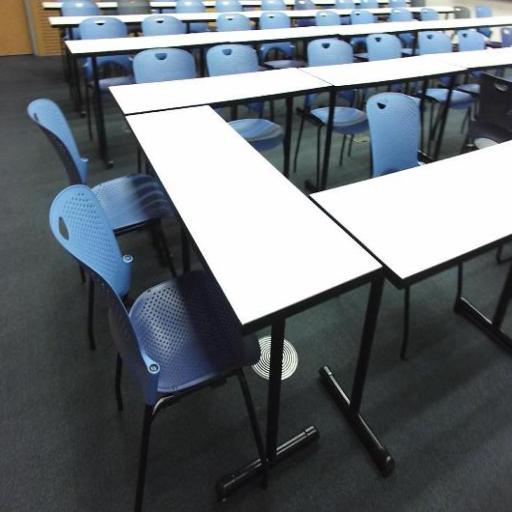}} 
  \label{fig:RGB543}{\includegraphics[width= \sizeOfimage\textwidth]{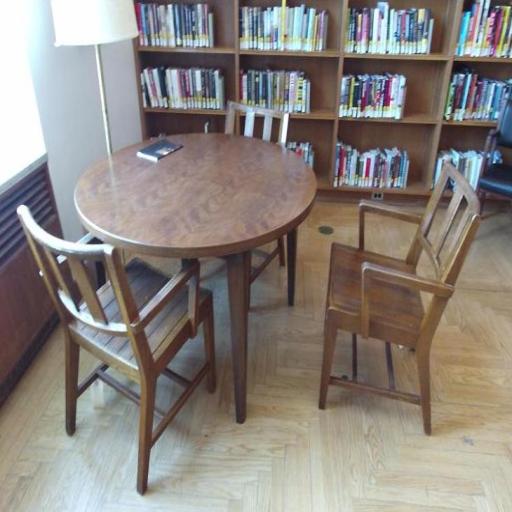}} 
  \label{fig:RGB1316}{\includegraphics[width= \sizeOfimage\textwidth]{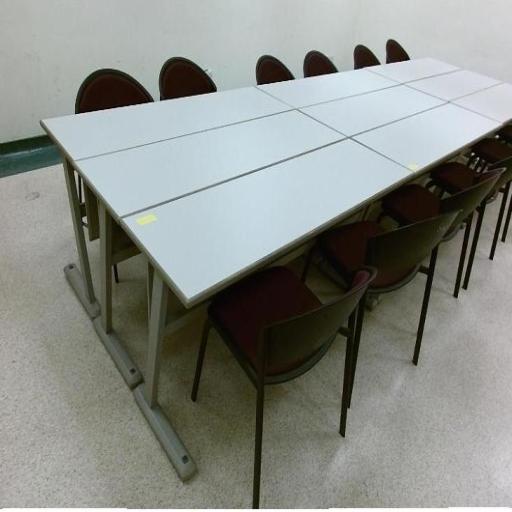}} 
  \label{fig:RGB1318}{\includegraphics[width= \sizeOfimage\textwidth]{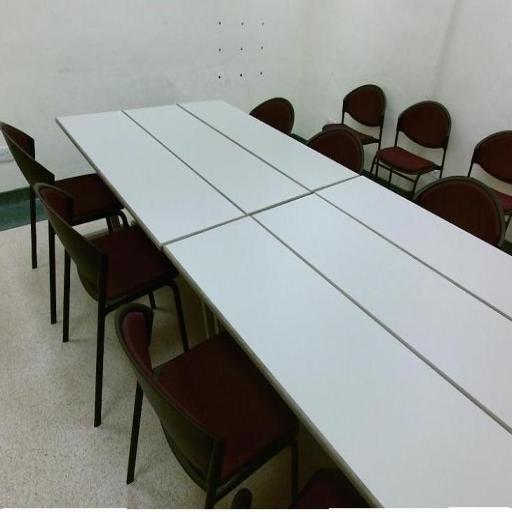}} 
  \label{fig:PointCloud39}{\includegraphics[width= \sizeOfimage\textwidth]{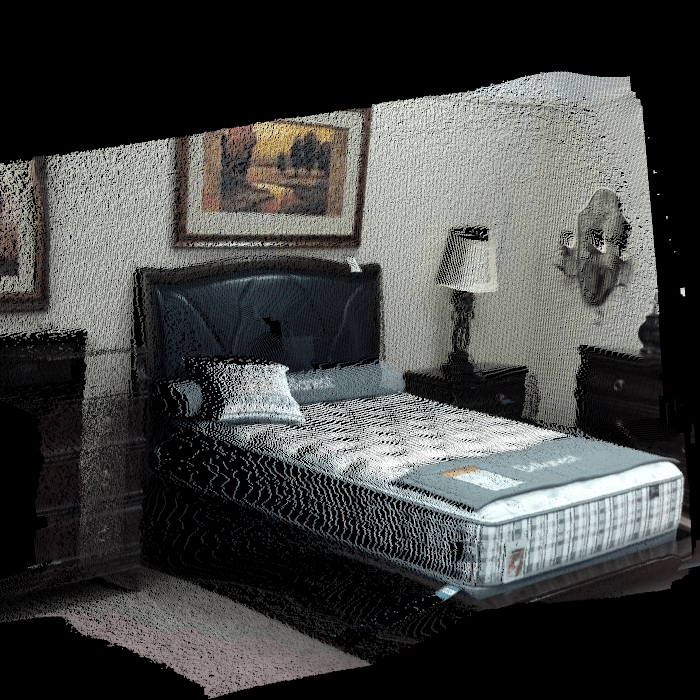}} 
  \label{fig:PointCloud453}{\includegraphics[width= \sizeOfimage\textwidth]{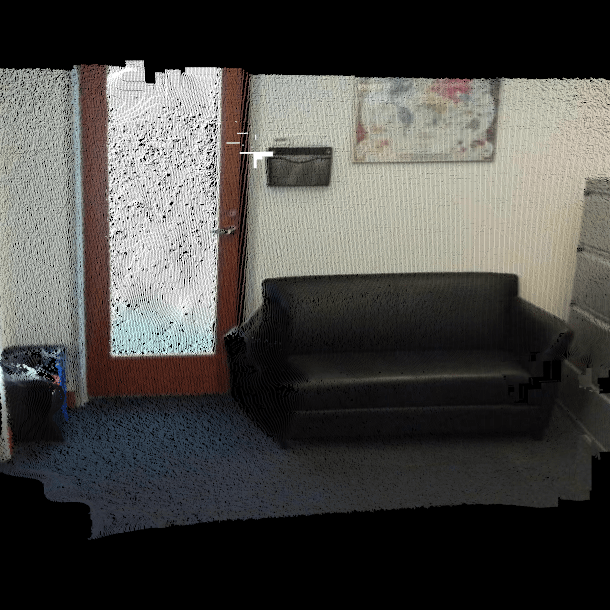}} 
  \label{fig:PointCloud4088}{\includegraphics[width= \sizeOfimage\textwidth]{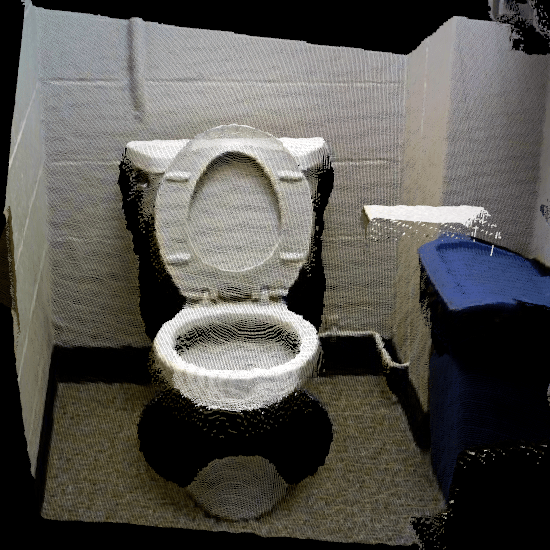}} 
  \label{fig:PointCloud4425}{\includegraphics[width= \sizeOfimage\textwidth]{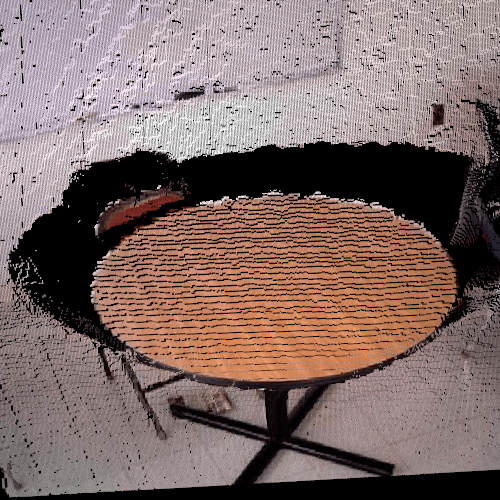}} 
  \label{fig:PointCloud162}{\includegraphics[width= \sizeOfimage\textwidth]{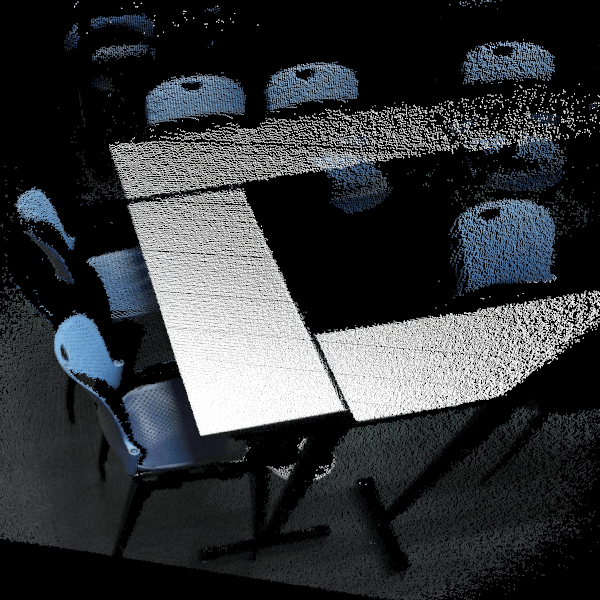}} 
  \label{fig:PointCloud543}{\includegraphics[width= \sizeOfimage\textwidth]{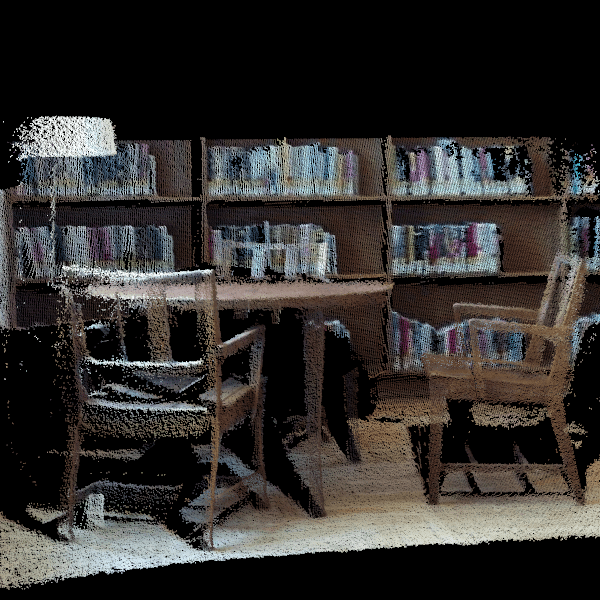}} 
  \label{fig:PointCloud1316}{\includegraphics[width= \sizeOfimage\textwidth]{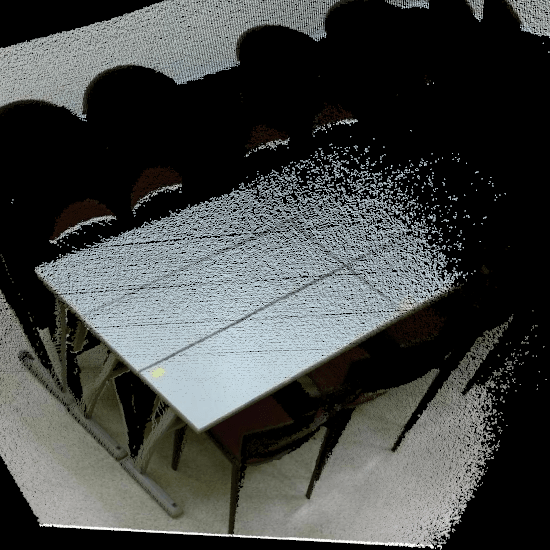}} 
  \label{fig:PointCloud1318}{\includegraphics[width= \sizeOfimage\textwidth]{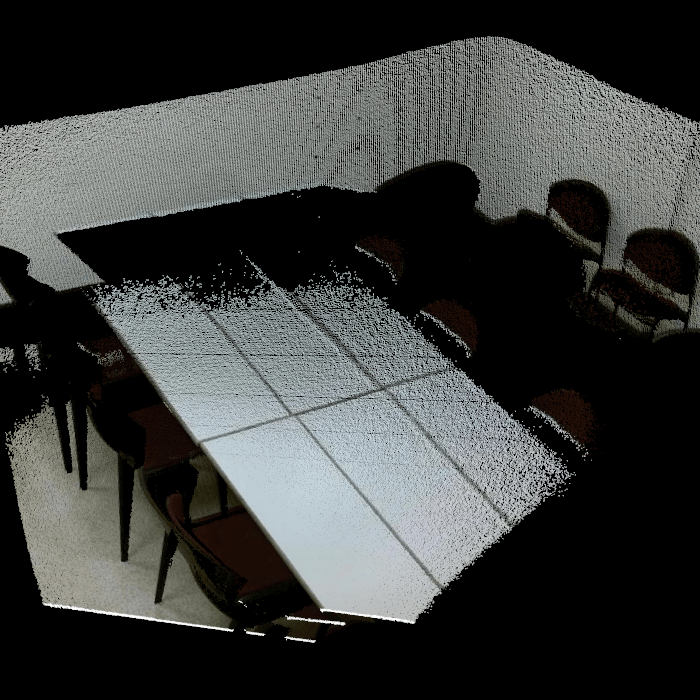}} 
  \label{fig:GT39}{\includegraphics[width= \sizeOfimage\textwidth]{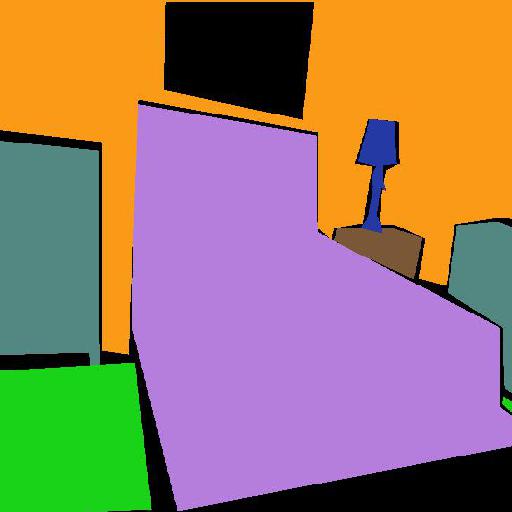}} 
  \label{fig:GT453}{\includegraphics[width= \sizeOfimage\textwidth]{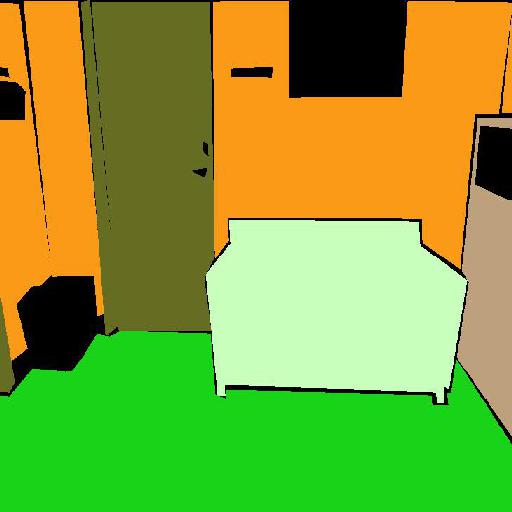}} 
  \label{fig:GT4088}{\includegraphics[width= \sizeOfimage\textwidth]{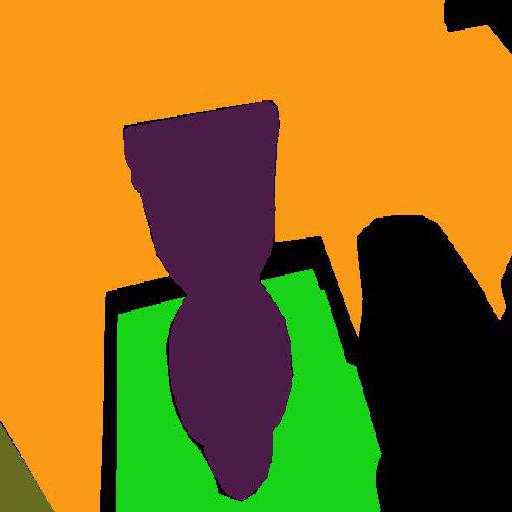}} 
  \label{fig:GT4425}{\includegraphics[width= \sizeOfimage\textwidth]{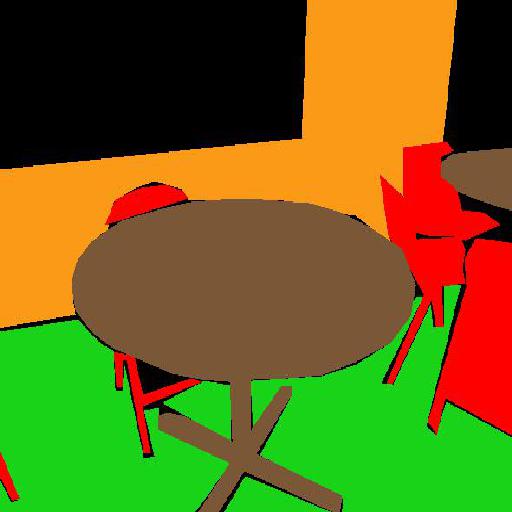}} 
  \label{fig:GT162}{\includegraphics[width= \sizeOfimage\textwidth]{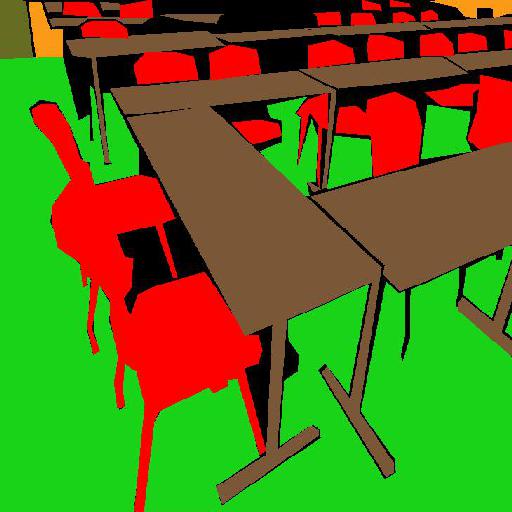}} 
  \label{fig:GT543}{\includegraphics[width= \sizeOfimage\textwidth]{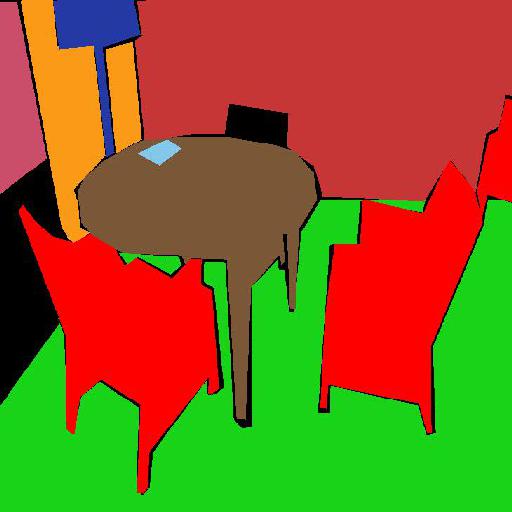}} 
  \label{fig:GT1316}{\includegraphics[width= \sizeOfimage\textwidth]{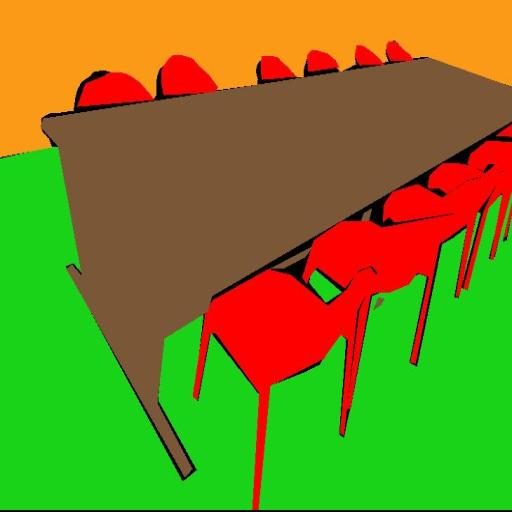}} 
  \label{fig:GT1318}{\includegraphics[width= \sizeOfimage\textwidth]{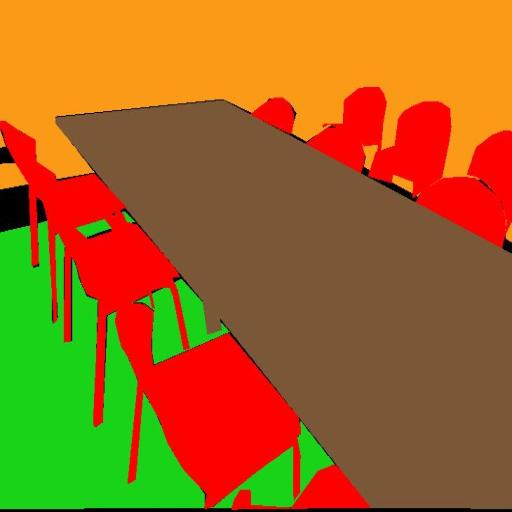}} 
  \label{fig:2D_prediction39}{\includegraphics[width= \sizeOfimage\textwidth]{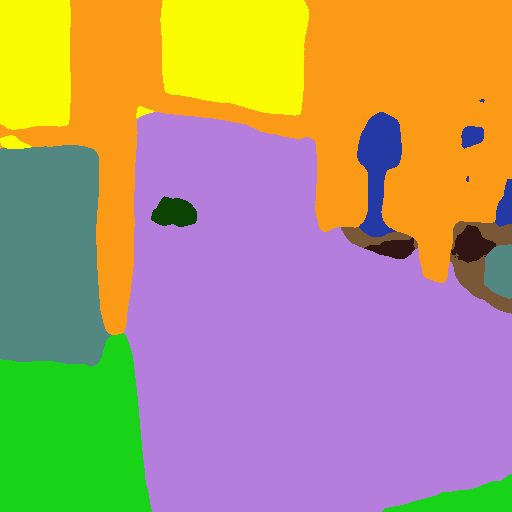}} 
  \label{fig:2D_prediction453}{\includegraphics[width= \sizeOfimage\textwidth]{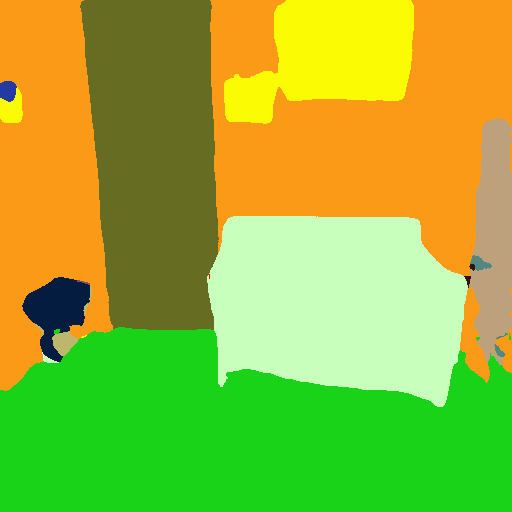}} 
  \label{fig:2D_prediction4088}{\includegraphics[width= \sizeOfimage\textwidth]{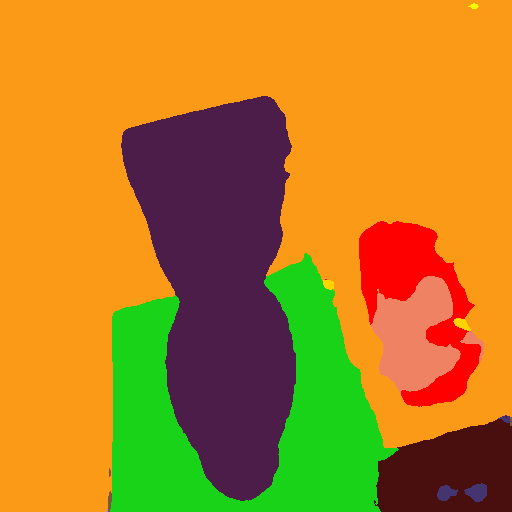}} 
  \label{fig:2D_prediction4425}{\includegraphics[width= \sizeOfimage\textwidth]{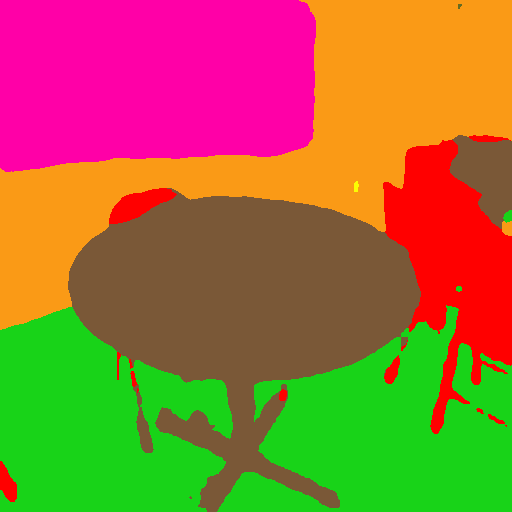}}
  \label{fig:2D_prediction162}{\includegraphics[width= \sizeOfimage\textwidth]{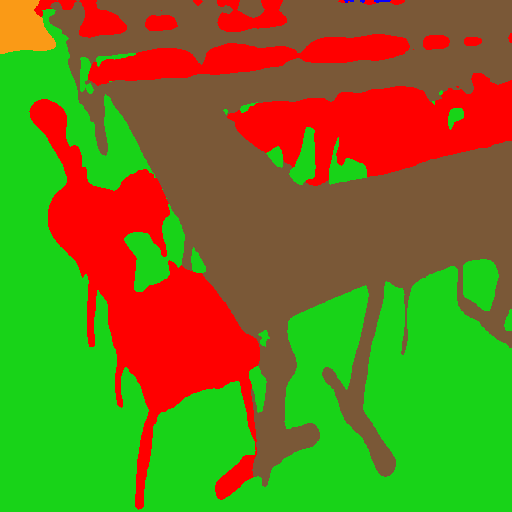}} 
  \label{fig:2D_prediction543}{\includegraphics[width= \sizeOfimage\textwidth]{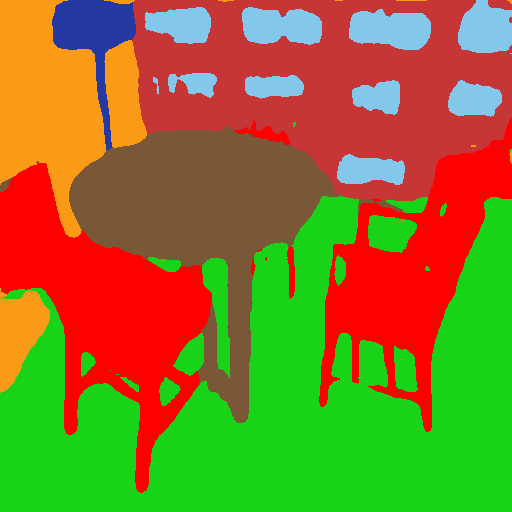}} 
  \label{fig:2D_prediction1316}{\includegraphics[width= \sizeOfimage\textwidth]{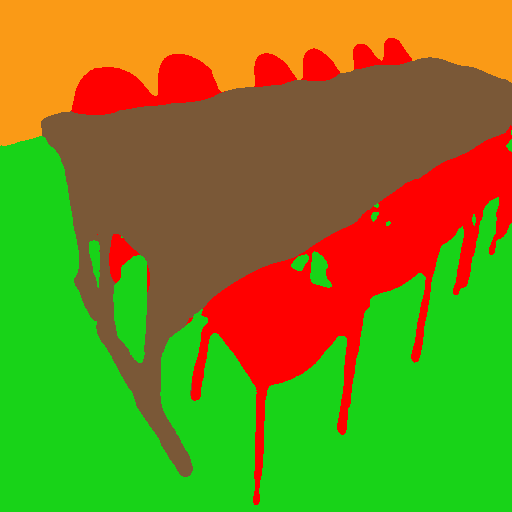}} 
  \label{fig:2D_prediction1318}{\includegraphics[width= \sizeOfimage\textwidth]{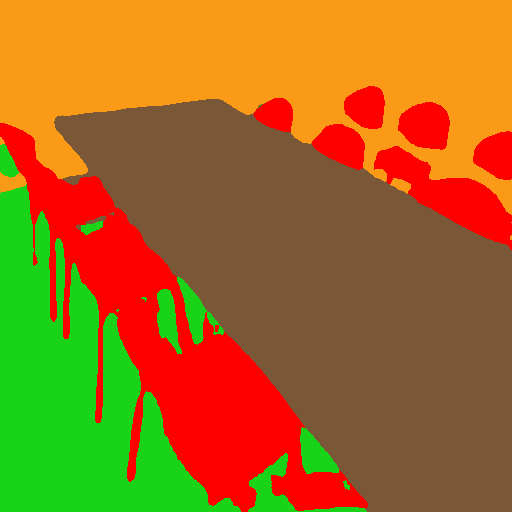}} 
  \label{fig:3D_prediction39}{\includegraphics[width= \sizeOfimage\textwidth]{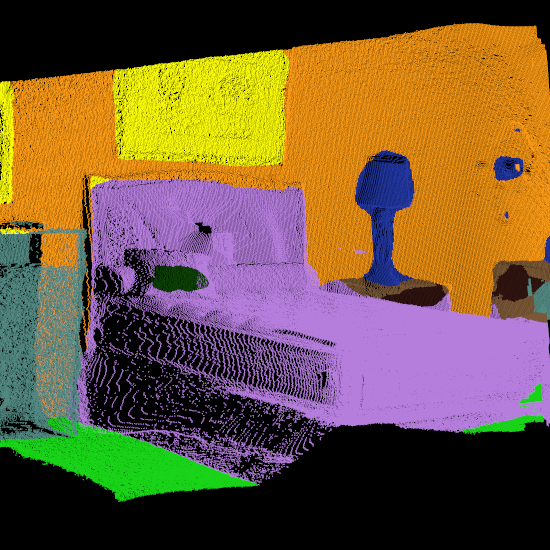}} 
  \label{fig:3D_prediction453}{\includegraphics[width= \sizeOfimage\textwidth]{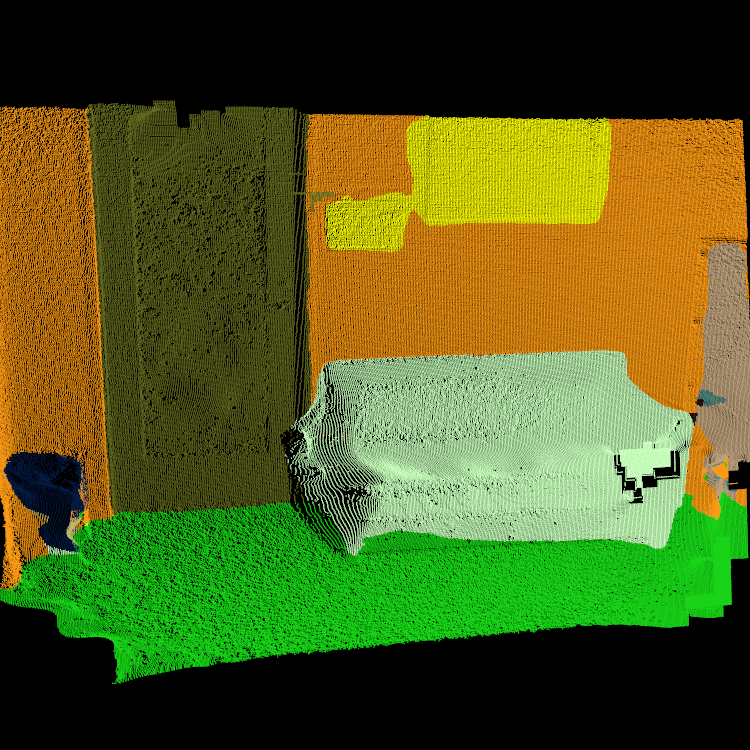}} 
  \label{fig:3D_prediction4088}{\includegraphics[width= \sizeOfimage\textwidth]{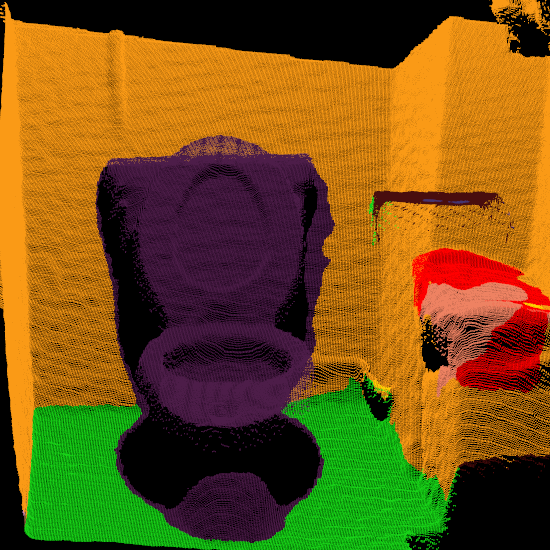}} 
  \label{fig:3D_prediction4425}{\includegraphics[width= \sizeOfimage\textwidth]{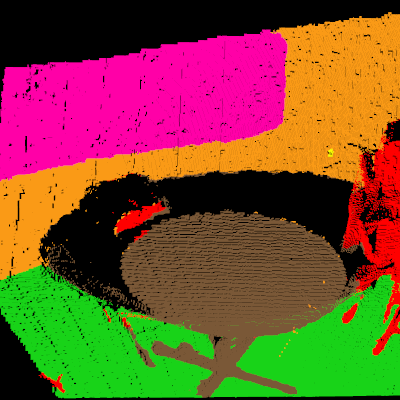}} 
  \label{fig:3D_prediction162}{\includegraphics[width= \sizeOfimage\textwidth]{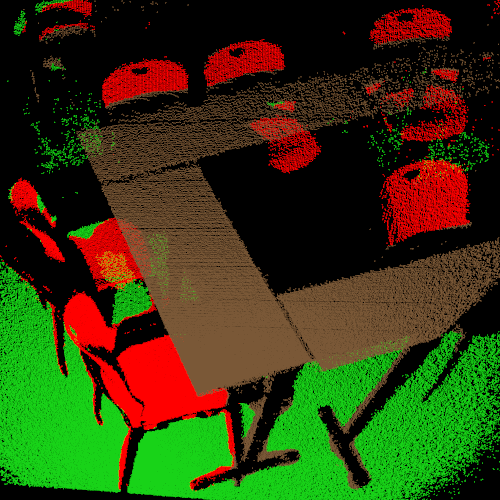}}
  \label{fig:3D_prediction543}{\includegraphics[width= \sizeOfimage\textwidth]{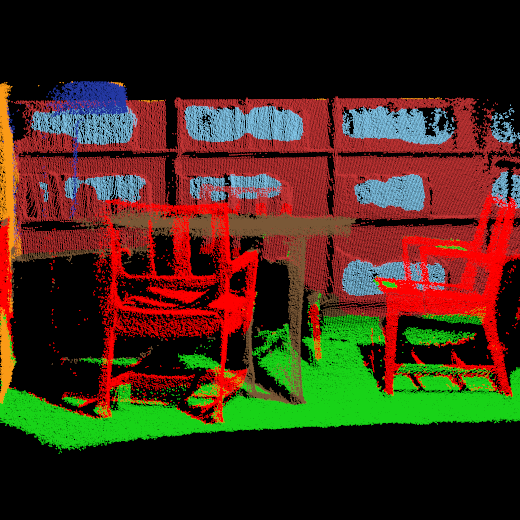}} 
  \label{fig:3D_prediction1316}{\includegraphics[width= \sizeOfimage\textwidth]{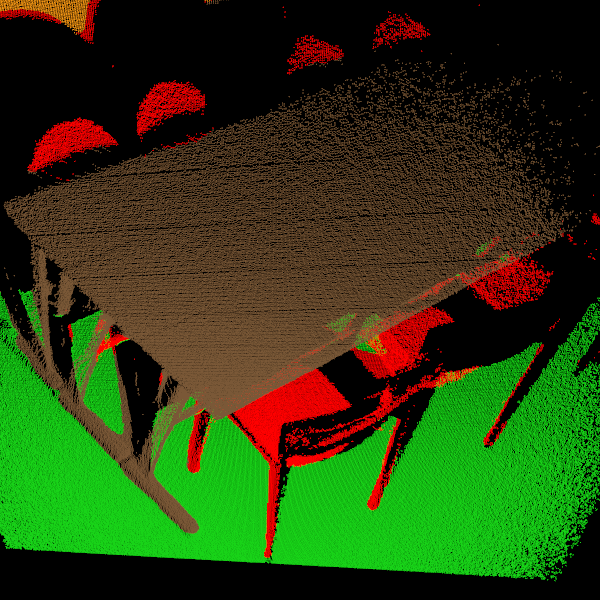}} 
  \label{fig:3D_prediction1318}{\includegraphics[width= \sizeOfimage\textwidth]{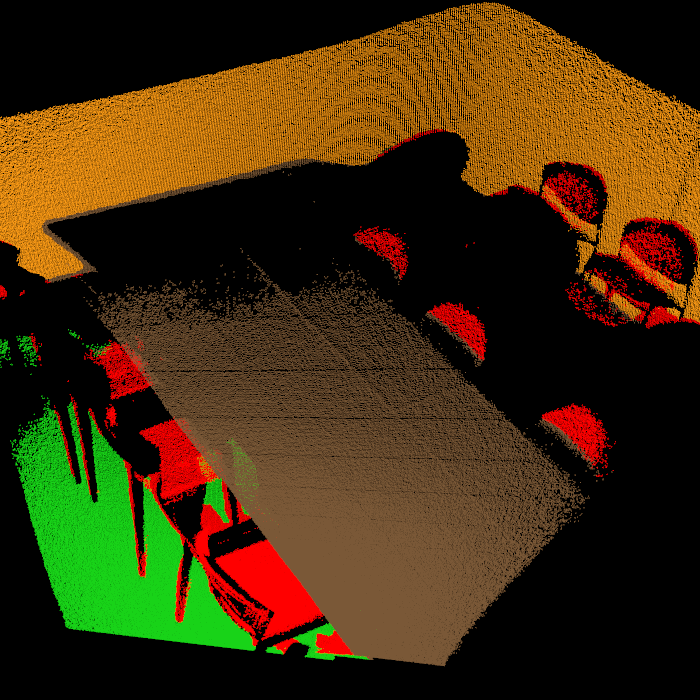}} 
  \label{fig:color_palette}{\includegraphics[width= 0.93\textwidth]{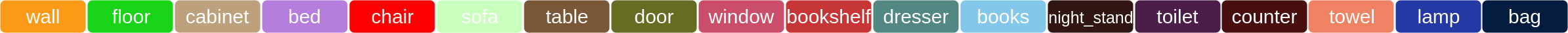}}

\caption{\textbf{The qualitative results (best viewed in colour) of Pixel-Voxel network on the SUN RGB-D dataset.} For different scenes in each row, the following images are displayed: RGB image(row 1), 3D point cloud(row 2), ground truth image(row 3), 2D semantic image(row 4) and 3D semantic point cloud(row 5). The Pixel-Voxel network produces the results with accurate boundary shape such as the shape of the bed, close-stool and especially the legs of furniture.}
\label{fig:qualitative_PVNet}
\end{figure*}

\begin{table}[thpb]
\caption{The comparison of \textbf{overall performance} on the SUN RGB-D dataset. Some reported results are copied from \cite{badrinarayanan2015segnet}.}
\centering
\resizebox{\columnwidth}{!}{
\begin{tabular}{ c | c  c  c }
\toprule
SUN RGB-D          & Pixel acc. & Mean acc. & Mean IoU\\
\midrule
FCN~\cite{long2015fully} & 68.18\% & 38.41\% & 27.39\% \\
DeconvNet~\cite{noh2015learning} & 66.13\% & 33.28\% & 22.57\% \\
SegNet~\cite{badrinarayanan2015segnet} & 72.63\% & 44.76\% & 31.84\% \\
DeepLab~\cite{chen2016deeplab} & 71.90\% & 42.21\% & 32.08\% \\
Context-CRF~\cite{lin2017exploring} & 78.4\% & 53.4\% & 42.3\% \\
LSTM-CF~\cite{li2016lstm}(RGB-D) & - & 48.1\% & - \\
FuseNet~\cite{hazirbas2016fusenet}(RGB-D) & 76.27\% & 48.30\% & 37.29\% \\
\midrule
Pixel-Voxel Net(VGG-16) & 78.14\% & 54.79\% & 42.11\% \\
Pixel-Voxel Net(ResNet101) & \textbf{79.04\%} & \textbf{57.65\%} & \textbf{44.24\%} \\
\bottomrule 
\end{tabular}}
\label{table:The comparison overall performance on the SUN RGB-D dataset.}
\end{table}

\begin{table*}[thpb]
\caption{The comparison of \textbf{class-wise accuracy} on the SUN RGB-D dataset. Not all the methods in Table \ref{table:The comparison overall performance on the SUN RGB-D dataset.} provide the class-wise accuracy in their papers. The \textbf{class-wise IoU} of Pixel-Voxel network (PVNet) is also provided.}
\centering
\resizebox{2\columnwidth}{!}{
\begin{tabular}{ c | c  c  c  c  c  c  c  c  c  c  c  c  c}
\toprule
Category                               & wall & floor & cabinet & bed & chair & sofa & table & door & window & bookshelf & picture & counter & blinds\\
\midrule
SegNet~\cite{badrinarayanan2015segnet}  & 83.42\% & 93.43\% & 63.37\% & 73.18\% & 75.92\% & 59.57\% & 64.18\% & 52.50\% & 57.51\% & 42.05\% & 56.17\% & 37.66\% & 40.29\% \\
LSTM-CF~\cite{li2016lstm}                  & 74.9\%  & 82.3\%  & 47.3\%  & 62.1\%  & 67.7\%  & 55.5\%  & 57.8\%  & 45.6\%  & 52.8\%  & 43.1\%  & 56.7\%  & 39.4\%  & 48.6\% \\
FuseNet~\cite{hazirbas2016fusenet}      & 90.20\% & 94.91\% & 61.81\% & 77.10\% & 78.62\% & 66.49\% & \textbf{65.44\%} & 46.51\% & 62.44\% & 34.94\% & \textbf{67.39\%} & 40.37\% & 43.48\% \\
PVNet(VGG16)                           & \textbf{90.28\%} & 93.21\% & 66.87\% & 75.31\% & 85.45\% & \textbf{67.37\%} & 64.81\% & 58.62\% & 63.58\% & 54.54\% & 64.76\% & \textbf{51.87\%} & \textbf{59.23\%}\\
PVNet(ResNet101)                       & 89.19\% & \textbf{94.94\%} & \textbf{69.36\%} & \textbf{79.11\%} & \textbf{85.70\%} & 66.09\% & 60.59\% & \textbf{62.22\%} & \textbf{66.59\%} & \textbf{58.34\%} & 66.39\% & 50.56\% & 53.65\%\\
\midrule
PVNet(VGG16)$_{IoU}$                   & 76.07\% & 87.20\% & 50.66\% & 68.23\% & 64.98\% & 54.17\% & 46.07\% & 44.83\% & 46.50\% & 41.31\% & 48.94\% & 41.19\% & 39.95\% \\
PVNet(ResNet101)$_{IoU}$               & 77.41\% & 87.78\% & 53.44\% & 71.16\% & 66.76\% & 54.61\% & 44.46\% & 45.19\% & 48.23\% & 41.79\% & 46.78\% & 41.39\% & 35.95\% \\
\toprule

Category                               & desk & shelves & curtain & dresser & pillow & mirror & floor\_mat & clothes & ceiling & books & fridge & tv & paper\\
\midrule
SegNet~\cite{badrinarayanan2015segnet}  & 11.92\% & 11.45\% & 66.56\% & 52.73\% & 43.80\% & 26.30\% & 0.00\% & 34.31\% & 74.11\% & 53.77\% & 29.85\% & 33.76\% & 22.73\% \\
LSTM-CF~\cite{li2016lstm}               & \textbf{37.3\%}  & 9.6\%   & 63.4\%  & 35.0\%  & 45.8\%  & 44.5\%  & 0.0\%  & 28.4\%  & 68.0\%  & \textbf{47.9\%}  & 61.5\%  & 52.1\%  & 36.4\% \\
FuseNet~\cite{hazirbas2016fusenet}      & 25.63\% & 20.28\% & 65.94\% & 44.03\% & 54.28\% & 52.47\% & 0.00\% & 25.89\% & \textbf{84.77\%} & 45.23\% & 34.52\% & 34.83\% & 24.08\% \\
PVNet(VGG16)                           & 32.05\% & 23.09\% & 62.49\% & 62.13\% & 54.97\% & 50.60\% & \textbf{0.59\%} & 35.35\% & 57.78\% & 41.75\% & 55.43\% & 67.60\% & 35.34\%\\
PVNet(ResNet101)                       & 32.49\% & \textbf{27.37\%} & \textbf{68.33\%} & \textbf{69.41\%} & \textbf{56.96\%} & \textbf{57.94\%} & 0.00\% & \textbf{36.45\%} & 68.77\% & 42.02\% & \textbf{63.05\%} & \textbf{72.47\%} & \textbf{38.11\%}\\
\midrule
PVNet(VGG16)$_{IoU}$                   & 26.05\% & 12.05\% & 50.52\% & 47.43\% & 36.35\% & 36.44\% & 0.59\%  & 20.56\% & 53.61\% & 28.04\% & 41.23\% & 57.36\% & 24.13\%\\
PVNet(ResNet101)$_{IoU}$               & 25.30\% & 16.86\% & 53.09\% & 50.83\% & 38.16\% & 42.29\% & 0.00\%  & 22.28\% & 63.39\% & 29.21\% & 48.47\% & 60.46\% & 25.20\% \\
\toprule

Category                               & towel & shower\_curtain & box & whiteboard & person & night\_stand & toilet & sink & lamp & bathtub & bag & mean & - \\
\midrule
SegNet~\cite{badrinarayanan2015segnet}  & 19.83\% & 0.03\% & 23.14\% & 60.25\% & 27.27\% & 29.88\% & 76.00\% & 58.10\% & 35.27\% & 48.86\% & 16.76\% & 31.84\% & - \\
LSTM-CF~\cite{li2016lstm}                  & 36.7\%  & 0.0\%  & 38.1\%  & 48.1\%  & \textbf{72.6\%}  & 36.4\%  & 68.8\%  & 67.9\%  & 58.0\%  & 65.6\%  & 23.6\%  & 48.1\%  & - \\
FuseNet~\cite{hazirbas2016fusenet}      & 21.05\% & \textbf{8.82\%} & 21.94\% & 57.45\% & 19.06\% & 37.15\% & 76.77\% & 68.11\% & 49.31\% & 73.23\% & 12.62\% & 48.30\% & - \\
PVNet(VGG16)                           & 41.12\% & 4.59\% & 40.33\% & 66.56\% & 60.51\% & 33.21\% & \textbf{80.62\%} & \textbf{69.07\%} & 60.35\% & 67.78\% & 28.17\% & 54.79\% & - \\
PVNet(ResNet101)                       & \textbf{48.81\%} & 0.00\% & \textbf{42.15\%} & \textbf{74.22\%} & 69.40\% & \textbf{38.16\%} & 80.23\% & 68.20\% & \textbf{61.80\%} & \textbf{76.16\%} & \textbf{37.63\%} & \textbf{57.65\%} & - \\
\midrule
PVNet(VGG16)$_{IoU}$                   & 30.53\% & 4.00\% & 24.81\% & 51.10\% & 48.57\% & 20.89\% & 66.31\% & 48.82\% & 43.50\% & 55.90\% & 19.37\% & 42.11\% & - \\
PVNet(ResNet101)$_{IoU}$               & 36.85\% & 0.00\% & 26.77\% & 54.88\% & 54.77\% & 21.52\% & 66.43\% & 53.15\% & 43.00\% & 65.00\% & 23.90\% & 44.24\% & - \\
\bottomrule
\end{tabular}}
\label{table:The comparison class-wise accuracy on the SUN RGB-D dataset.}
\end{table*}

\begin{figure*}[thpb]
\centering
\label{fig:PointCloud}{\includegraphics[width= 0.8\textwidth]{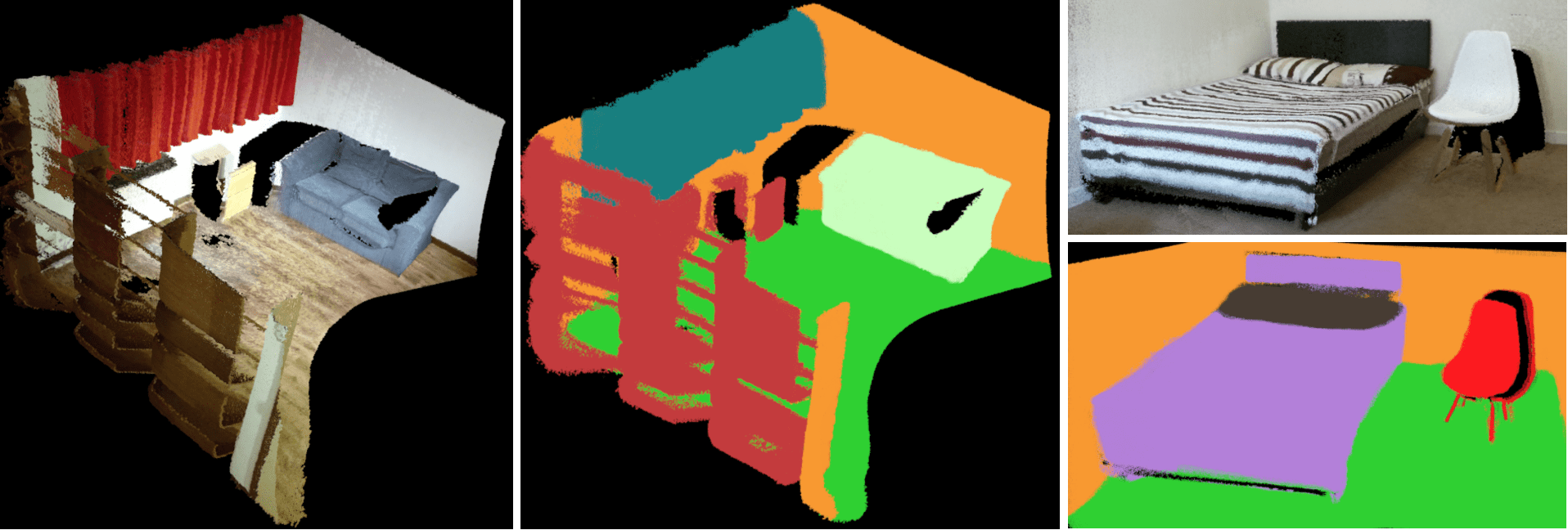}}

\label{fig:map_palette}{\includegraphics[width= 0.6\textwidth]{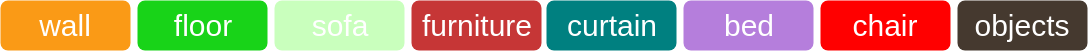}}
\caption{The \textbf{dense 3D map} and \textbf{dense 3D semantic map} (best viewed in colour) of a living room and bedroom.}
\label{fig:semantic mapping}
\end{figure*}

\subsection{Dense RGB-D semantic mapping}\label{sec:4.4}
The dense RGB-D semantic mapping system is implemented under the ROS\footnote{\url{http://www.ros.org/}} framework and runs on a desktop with i7-6800k ($3.4Hz$) 8-cores CPU and NVIDIA TITAN X GPU (12G). Kinect V2 is used to obtain the RGB images and point clouds. IAI Kinect2 package2\footnote{\url{https://github.com/code-iai/iai kinect2/}} is employed to interface with ROS and calibrate the Kinect2 cameras. The Pixel-Voxel network is implemented using Caffe\footnote{\url{http://caffe.berkeleyvision.org/}} toolbox. It is trained on the TITAN X GPU, accelerated by CUDA and CUDNN.

The system with a pre-trained network is tested in the real-world environment, i.e., a living room and bedroom containing the curtain, bed and etc., as shown in Figure \ref{fig:semantic mapping}. It can be seen that most of the voxels are correctly segmented and the results have accurate boundary shapes. But there are still some voxels in the boundary to be assigned wrong predictions. Some error predictions are caused by upsampling the data through a bilateral filter to the same size as Kinect V2 data. Another reason is that this network is trained using the public SUN RGB-D dataset but it is tested using the real-world data. So some errors result from illumination variances, categories variances and etc. In addition, the noise of Kinect V2 also causes some error predictions.   

The runtime performance of our system is $5-6Hz$ using the QHD data from Kinect2. During real-time RGB-D mapping, only a few key-frames are used for mapping. Most of the frames are abandoned because of the small variance between two consecutive frames. It is not necessary to segment all the frames in the sequence but only the key-frames. As mentioned in~\cite{hermans2014dense}, $5Hz$ runtime performance can nearly satisfy the real-time dense 3D semantic mapping. The runtime performance can be boosted to $11-12Hz$ using the half scale data. It is a trade-off between runtime and accuracy.

\textbf{All the source code will be published upon acceptance of this paper. A real-time demo can be found in this link \url{https://youtu.be/UbmfGsAHszc}.}

\section{Conclusion}\label{sec:5}
In this paper, a dense RGB-D semantic mapping system is developed for the real-time applications. The runtime of the system can be boosted to $11-12Hz$ using an i7 8-cores PC with Titan X GPU. A  Pixel-Voxel network is proposed that achieves the state-of-the-art semantic segmentation performance on SUN RGB-D benchmark dataset. The proposed Pixel-Voxel Network integrates: 1) PixelNet that aggregates the multi-scales global context information from the RGB image, which extends the receptive field to cover all the elements in the feature map by utilizing multiple context stacks. 2) VoxelNet that preserves the local shape information of the 3D point cloud under the absence of conventional pooling layer. We also proposed a Softmax weighted fusion stack that combines PixelNet and VoxelNet together according to their respective confidence levels under different situations. The qualitative and quantitative evaluations on SUN RGB-D dataset and real-world datasets confirm the effectiveness of the proposed Pixel-Voxel Network. 

\section{Acknowledgement}
The work was supported by Toshiba Research Europe and DISTINCTIVE scholarship.
\bibliographystyle{IEEEtran}
\bibliography{./references_short.bib}
\end{document}